\documentclass[preprint,12pt]{elsarticle}

\usepackage{amssymb}
\usepackage{amsmath}
\usepackage{amsmath,amssymb,amsfonts}
\usepackage{algorithmic}
\usepackage{graphicx}
\usepackage{textcomp}
\usepackage{booktabs}
\usepackage{hyperref}
\usepackage{color}
\usepackage{xcolor}
\usepackage{multirow}
\def\etal{\textit{et al.}}

\journal{journal}

\begin{document}

\begin{frontmatter}



\title{Facial Emotion Learning with Text-Guided Multiview Fusion via Vision-Language Model for 3D/4D Facial Expression Recognition}

\author[1,2]{Muzammil Behzad\corref{cor1}}
\ead{muzammil.behzad@kfupm.edu.sa}

\cortext[cor1]{Corresponding author}

\affiliation[1]{organization={King Fahd University of Petroleum and Minerals},
            country={Saudi Arabia}}
\affiliation[2]{organization={SDAIA-KFUPM Joint Research Center on Artificial Intelligence},
            country={Saudi Arabia}} 

\begin{abstract}
Facial expression recognition (FER) in 3D and 4D domains presents a significant challenge in affective computing due to the complexity of spatial and temporal facial dynamics. Its success is crucial for advancing applications in human behavior understanding, healthcare monitoring, and human-computer interaction. In this work, we propose FACET-VLM, a vision-language framework for 3D/4D FER that integrates multiview facial representation learning with semantic guidance from natural language prompts. FACET-VLM introduces three key components: Cross-View Semantic Aggregation (CVSA) for view-consistent fusion, Multiview Text-Guided Fusion (MTGF) for semantically aligned facial emotions, and a multiview consistency loss to enforce structural coherence across views. Our model achieves state-of-the-art accuracy across multiple benchmarks, including BU-3DFE, Bosphorus, BU-4DFE, and BP4D-Spontaneous. We further extend FACET-VLM to 4D micro-expression recognition (MER) on the 4DME dataset, demonstrating strong performance in capturing subtle, short-lived emotional cues. The extensive experimental results confirm the effectiveness and substantial contributions of each individual component within the framework. Overall, FACET-VLM offers a robust, extensible, and high-performing solution for multimodal FER in both posed and spontaneous settings.
\end{abstract}




\end{frontmatter}




\section{Introduction}

Recent advancements in vision-language models (VLMs) have significantly transformed the landscape of artificial intelligence by enabling an in-depth understanding across visual and linguistic modalities~\cite{bordes2024introductionvisionlanguagemodeling}. Derived from the fundamental architecture of large language models (LLMs)~\cite{minaee2024largelanguagemodelssurvey}, these models are trained on massive multimodal datasets to jointly embed images and text in a shared representation space. One of the most successful instantiations of this paradigm is the contrastive language-image pretraining (CLIP)~\cite{radford2021learningtransferablevisualmodels} model, which uses contrastive objectives to align image and text pairs, enabling a broad range of downstream tasks including open-vocabulary classification and zero-shot learning. As a result, VLMs have rapidly become foundational components in multimodal tasks, particularly because they not only learn robust semantic priors from complicted data but also transfer efficiently to new domains~\cite{zhai2024finetuninglargevisionlanguagemodels}.

Some of the most important applications of VLMs include facial expression analysis. This is because facial expression recognition (FER) continues to play a central role in affective computing. The ability pf these models to infer emotional states from facial cues offers a variety of applications ranging from emotion-aware virtual agents and adaptive tutoring systems~\cite{YADEGARIDEHKORDI2019103649} to mental health monitoring~\cite{Foteinopoulou_2022} and human behavior modeling~\cite{7374704}. In this regard, traditional FER approaches have largely been built upon static 2D facial images and manual feature engineering~\cite{LIU2023423}, based on Ekman’s pioneering theory of six basic emotions~\cite{ekman1971constants}. While these approaches are effective in constrained environments, they struggle to generalize to unconstrained, in-the-wild scenarios, particularly when dealing with variations in pose, lighting, expression intensity, and identity.

To overcome the limitations of 2D-based facial expression recognition, the field has steadily progressed toward 3D and 4D facial expression analysis. These modalities enrich the input space with depth (3D) and motion (4D), enabling more accurate modeling of subtle muscle deformations and facial dynamics. In static 3D analysis, a variety of approaches have been developed to exploit facial surface geometry. For instance, local geometric descriptors~\cite{6460694, 5206613, li2015efficient} have been employed to extract curvature-based or point-level features across facial landmarks and regions. Similarly, template-based methods~\cite{4539275, 5597896} align raw 3D scans with reference meshes, enabling deformation measurement through distance maps or flow fields. Likewise, shape-aware descriptors, including curve-based encodings~\cite{samir2009intrinsic, maalej2011shape}, analyze intrinsic surface properties by tracing geodesic lines or surface normals to model regional deformations. These early systems effectively capture identity-normalized features but often require carefully tuned spatial priors and are sensitive to mesh resolution and noise.

To bridge the gap between traditional 3D processing and modern deep learning pipelines, projection-based techniques~\cite{7944639, 8265585} have been proposed to transform 3D meshes into 2D maps (such as, depth, curvature, or normal images), so that convolutional neural networks can be applied. On the temporal side, 4D FER leverages the full sequence of mesh frames to model expression evolution. In this context, the probabilistic models like Hidden Markov Models (HMMs)~\cite{4813324, Sun:2010:TVF:1820799.1820803} capture the stochastic progression of expressions, while ensemble classifiers such as GentleBoost~\cite{sandbach2012recognition} and deformation-based forest models~\cite{amor20144} incorporate motion patterns between frames. Spatiotemporal encoders such as LBP-TOP~\cite{FANG2012738, 6130440} extend the success of 2D texture features to 4D by capturing appearance and motion from orthogonal planes. Additionally, dynamic curvature-based representations~\cite{6553746} leverage local surface variation over time. Although these handcrafted and shallow-learned approaches laid critical groundwork for 4D FER, they are limited by their reliance on manually designed features and often require large quantities of aligned, high-quality 3D sequences. This motivates the exploration of more adaptive, scalable learning strategies capable of jointly leveraging multiview visual information and high-level semantic guidance. Building on these foundations, Li \etal~\cite{8373807} introduced a score-level fusion mechanism over multiple geometric projections from differential 3D data to enhance the robustness of 4D FER systems. This work has solidified the view that 3D and 4D modalities are inherently better suited for emotion analysis due to their richer structural content.

\subsection{Motivation}

Despite these advancements, there remains a fundamental disconnect between the semantic interpretation of expressions and the features used to model them. Most existing 3D/4D FER models still operate purely within the visual domain, extracting low-level geometric features or learning view-invariant embeddings without incorporating higher-level semantic priors. The success of VLMs in bridging vision and language provides a compelling opportunity to address this gap. By introducing text-based prompts as additional supervision signals, VLMs can semantically organize the learned visual space around meaningful affective concepts. For instance, natural language descriptors like ``a fearful face'' or ``a joyful smile'' can guide the network to associate specific geometric and motion patterns with human-interpretable emotion categories.

As a result, this paper introduces a new approach to 3D/4D FER that capitalizes on the joint representational strength of multiview vision and textual language. Our strategy centers on decomposing 3D or 4D facial data into three projected 2D views (frontal, left, and right) and processing these with a shared vision encoder. These views are semantically learned by aligning their features with language prompts in a shared embedding space via a pre-trained VLM. While previous 2D FER methods~\cite{8245803, 8844064, 9226082, 9369001} often struggle with generalization to different poses or subjects, our multiview approach improves robustness by aggregating pose-complementary information. Furthermore, the integration of text introduces semantic structure into the training objective, which facilitates more transferable emotion representations.

\begin{figure}[t!]
    \centering
    \includegraphics[width=\linewidth]{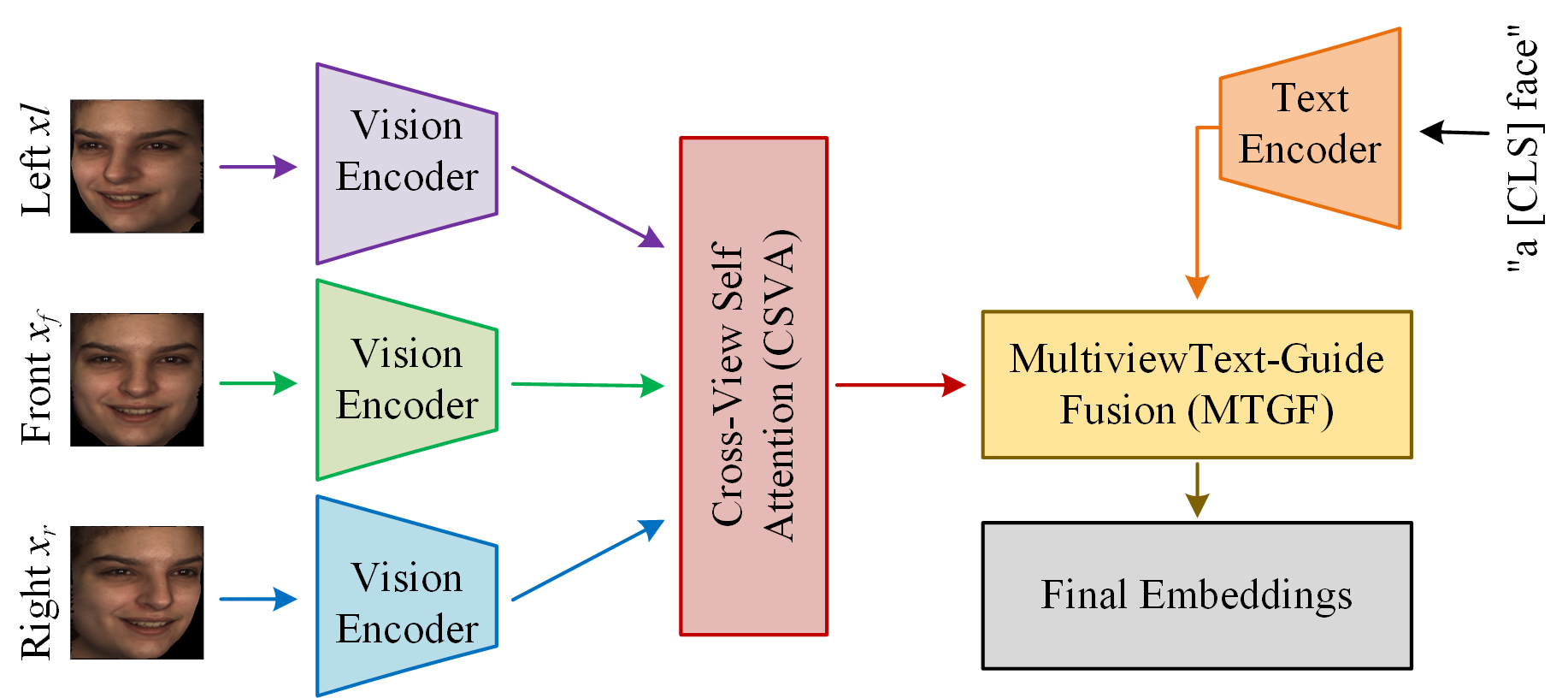}
    \caption{A brief overview of the proposed FACET-VLM architecture.}
    \label{fig:intro_model}
\end{figure}
\subsection{Multiview Vision-Language Modeling for 3D/4D FER}
In this work, we propose a unified FACET-VLM framework for 3D/4D facial expression recognition using vision-language modeling. As shown in Fig. \ref{fig:intro_model}, our approach leverages three core design principles. First, we employ a cross-view self-attention mechanism that facilitates feature interaction across multiple perspectives, enabling the model to synthesize holistic facial geometry from partial views. Second, we introduce a novel text-guided fusion module that conditions the visual fusion process on semantic information from natural language prompts. This enhances the discriminative power of the learned embeddings by tightly coupling visual cues with their corresponding emotional descriptors. Finally, we incorporate a consistency loss that enforces alignment between view-specific representations, ensuring stability and robustness across viewpoint variations.

These architectural choices are coupled with a multimodal training strategy that aligns visual tokens from the three views with textual embeddings in a shared space using a contrastive loss. Unlike previous methods that rely solely on categorical emotion labels, our system benefits from the compositional richness of natural language, enabling a more detailed understanding of facial affects. This results in a model that is robust to pose, interpretable via language, and effective even in settings with sparse supervision.

Consequently, FACET-VLM offers a new paradigm for facial expression recognition that integrates multiview geometry with semantic understanding through vision-language alignment. By bridging the gap between structured 3D/4D visual data and descriptive language cues, our method sets the stage for more scalable, human-centric emotion recognition systems in affective computing and beyond.

\subsection{Contributions}

In this work, we present FACET-VLM: Facial Emotion Learning with Text-Guided Multiview Fusion via Vision-Language Model for 3D/4D Facial Expression Recognition. FACET-VLM is designed to bridge the gap between geometric multiview facial representation and semantic language supervision in the context of 3D/4D facial expression recognition. It introduces a new perspective on how vision-language modeling can be leveraged for emotion understanding from 3D multiview projections of facial scans. Our main contributions are summarized as follows:

\begin{itemize}
    \item We propose a vision-language framework for 3D/4D FER that aligns multiview 2D projections of 3D facial data with natural language emotion prompts using a shared embedding space, enabling semantic understanding and cross-modal representation learning.
    
    \item We introduce a cross-view self-attention (CVSA) module that jointly attends over patch tokens from multiple views (front, left, right), enabling the model to aggregate discriminative facial cues across poses and produce view-complementary features.
    
    \item We design a novel multiview text-guided fusion (MTGF) layer that incorporates text supervision directly at the token-fusion stage through cross-attention, allowing emotion descriptions to modulate the visual fusion process for fine-grained affective discrimination.
    
    \item We incorporate a multiview consistency loss to regularize the learned embeddings across different views of the same expression, improving viewpoint invariance, stability, and generalization to unseen identities or camera angles.
    
    \item We conduct extensive evaluations on 3D/4D facial datasets and demonstrate that FACET-VLM achieves competitive results.
\end{itemize}

Overall, FACET-VLM offers a unified and semantically aligned approach for multiview 3D/4D facial expression recognition. By integrating natural language prompts with geometric visual data, our model enables flexible and data-efficient emotion understanding that generalizes across viewpoints and identities.

\section{The Proposed FACET-VLM Framework}
\begin{figure*}[t!]
    \centering
    \includegraphics[width=\linewidth]{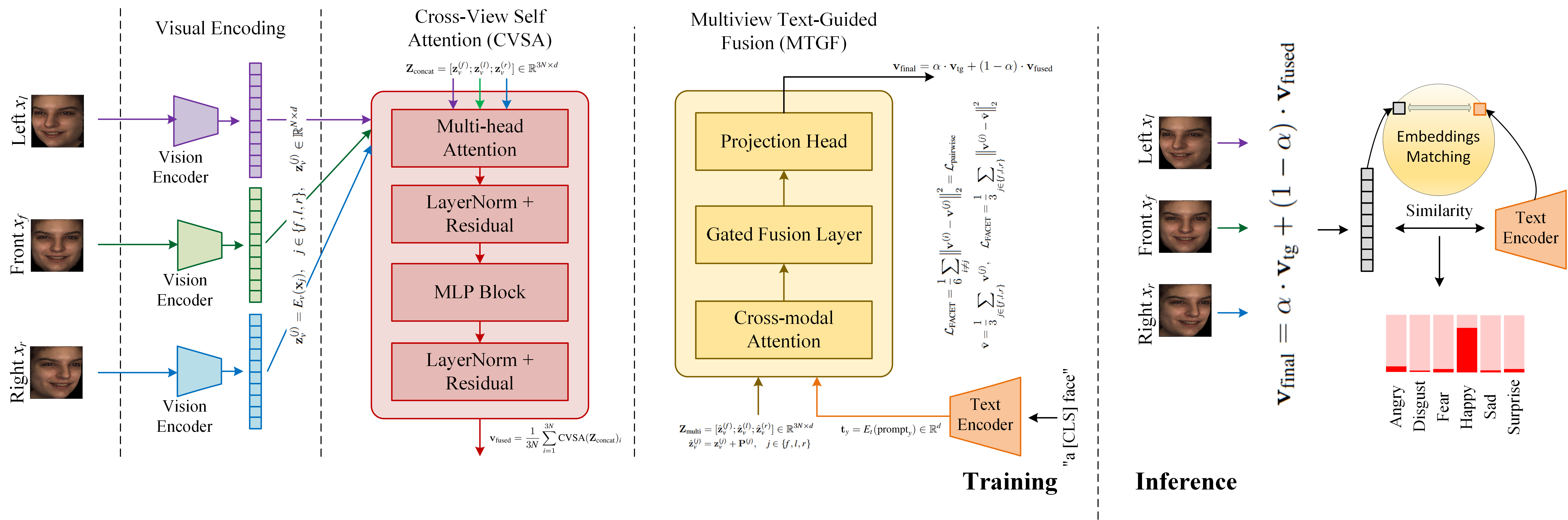}
    \caption{Overview of the FACET-VLM architecture for 3D/4D facial expression recognition. Multiview facial inputs (left, front, right) are independently processed through a shared Vision Encoder to produce view-specific token sequences. These tokens are concatenated and refined via the Cross-View Self Attention (CVSA) module, followed by Multiview Text-Guided Fusion (MTGF), where semantic guidance from a language prompt modulates the visual features through cross-modal attention and gated fusion. The final embedding is aligned with the text embedding using a contrastive loss during training and compared against expression prompts at inference time via similarity matching.}
    \label{fig:facet_vlm_architecture}
\end{figure*}

In this paper, we use CLIP~\cite{radford2021learningtransferablevisualmodels} as our baseline model to propose a multiview vision-language framework for 3D emotion recognition that leverages three facial views (front, left, right) for cross-modal representation learning. The pipeline consists of a multiview vision encoder with a novel cross-view self-attention mechanism, a language-guided fusion layer, and a proposed regularization loss to enforce consistency across views. An overview of the full FACET-VLM architecture, including training and inference components, is illustrated in Figure~\ref{fig:facet_vlm_architecture}.

\subsection{Multiview Emotion Dataset Preprocessing}

In this section, we outline the preprocessing pipeline required to transform raw 3D/4D facial emotion datasets into a format suitable for our multiview vision-language recognition framework. The preprocessing involves three key steps: (i) multiview image normalization to ensure geometric consistency, (ii) textual prompt engineering to formulate semantically rich emotion descriptions, and (iii) tokenization of visual data into patch-level embeddings using a shared vision encoder.

\subsubsection{Image Normalization}
To enable effective learning of correspondences across views, we construct a spatially normalized and temporally synchronized multiview data. Specifically, each 3D/4D facial instance is decomposed into three 2D image projections: a frontal view ($\mathbf{x}_f$), a left-profile view ($\mathbf{x}_l$), and a right-profile view ($\mathbf{x}_r$). Similarly, all images are resized to a standardized resolution $H \times W \times C$ prior to feeding into the encoder. This uniformity ensures compatibility with pretrained models, which typically expect fixed-size inputs (e.g., $224 \times 224$). Further, the color channels are normalized using mean and standard deviation statistics derived from either ImageNet or dataset-specific distributions, which accelerates convergence and ensures stable gradient propagation.

\subsubsection{Text Prompt Engineering}
Unlike traditional classification settings that use one-hot labels, our approach embeds the emotion category into natural language prompts that provide semantic richness and contextual cues. This is crucial in the context of contrastive learning using vision-language models which works in a joint image-text embedding space. Each label $y \in \mathcal{Y} = \{\texttt{happy}, \texttt{anger}, \texttt{disgust}, \\ \texttt{fear}, \texttt{sad}, \texttt{surprise}\}$ is mapped to a sentence prompt using a synthetic linguistic template. This template is designed to be descriptive yet compact, e.g., the label \texttt{happy} is converted to the prompt phrase \texttt{"a happy face"}. These prompts are passed to the text encoder $E_t(\cdot)$ to produce dense semantic embeddings. The design of such prompts is inspired by recent work in prompt engineering~\cite{minaee2024largelanguagemodelssurvey}, which has been shown to significantly influence zero-shot generalization and cross-modal alignment. A complete mapping of emotion labels to their respective prompts is provided in Table~\ref{tab:prompt-engineering}. This allows the model to align its visual understanding in emotionally descriptive language and facilitates better generalization to semantically similar but unseen expressions.

\begin{table}[b!]
\centering
\caption{Prompt generation for emotion labels.}
\label{tab:prompt-engineering}
\begin{tabular}{cc} 
\toprule 
\textbf{Emotion Label} & \textbf{Generated Prompt} \\
\midrule 
\texttt{happy} & "a happy face" \\
\texttt{anger} & "an angry face" \\
\texttt{disgust} & "a disgusted face" \\
\texttt{fear} & "a fearful face" \\
\texttt{sad} & "a sad face" \\
\texttt{surprise} & "a surprised face" \\
\bottomrule 
\end{tabular}
\end{table}

\subsubsection{Tokenization}
After preprocessing, each normalized image $\mathbf{x}_j$ for $j \in \{f, l, r\}$ is passed through a shared visual tokenizer based on the ViT-B/16 architecture. This encoder, denoted as $E_v(\cdot)$, first divides the image into non-overlapping patches of fixed size $p \times p$ (e.g., $16 \times 16$), resulting in $N = \frac{H \cdot W}{p^2}$ image tokens. Each patch is then linearly projected into a $d$-dimensional embedding, followed by position encoding and a series of self-attention transformer layers. The final tokenized representation is given by:
\begin{equation}
    \mathbf{z}_v^{(j)} = E_v(\mathbf{x}_j) \in \mathbb{R}^{N \times d}.
\end{equation}
These patch-level embeddings preserve local spatial structure which are then used in the subsequent multiview fusion and attention modules. Importantly, the encoder $E_v(\cdot)$ is shared across all views to promote weight sharing and parameter efficiency. This also ensures that the embeddings across views are present in the same latent space, thereby facilitating cross-view attention and alignment. 

This multiview preprocessing strategy forms the backbone for the subsequent components of our framework. The careful design choices in geometric normalization, semantic prompt generation, and consistent tokenization ensure that the model is well-positioned to learn robust and generalizable emotion representations across multiple viewpoints. By explicitly anchoring each view and emotion category into the same latent space, we enable effective fusion and comparison across modalities which is a key requirement for vision-language contrastive learning.

\subsection{Multiview Vision Encoder with Cross-View Attention}
While each view is initially processed independently by a vision encoder, our novel Cross-View Self-Attention (CVSA) mechanism facilitates direct inter-view communication at the token level. The resulting fused representation captures both spatial and cross-view semantic correspondences, enabling viewpoint-robust emotion classification.

\subsubsection{Shared Encoder Backbone}
We adopt the ViT-B/16 architecture from the CLIP model as the visual backbone $E_v(\cdot)$ for each view. This is because ViT-based encoders have demonstrated strong transfer performance due to their attention-based global context modeling and compatibility with language embeddings. To maintain parameter efficiency and shared feature space, the same encoder is applied to all three views. The encoder outputs patch-wise token sequences as:
\begin{equation}
    \mathbf{z}_v^{(j)} = E_v(\mathbf{x}_j), \quad j \in \{f, l, r\}, \quad \mathbf{z}_v^{(j)} \in \mathbb{R}^{N \times d}.
\end{equation}
These embeddings retain fine-grained spatial information and serve as the basis for inter-view fusion. During early training epochs, the encoder weights are frozen to preserve pretrained semantics. Subsequently, we unfreeze higher transformer layers to adapt to the emotion recognition domain.

\subsubsection{Cross-View Self-Attention}
To learn rich correspondences across views, we propose a Cross-View Self-Attention (CVSA) mechanism. Unlike standard ViTs which operate on a single image, CVSA receives tokens from all three views and jointly models their interactions. Specifically, we concatenate the tokens across views:
\begin{equation}
\mathbf{Z}_{\text{concat}} = [\mathbf{z}_v^{(f)}; \mathbf{z}_v^{(l)}; \mathbf{z}_v^{(r)}] \in \mathbb{R}^{3N \times d}.
\end{equation}
This allows attention heads to attend across spatial locations and across viewpoints, enabling cross-view semantic reasoning. The joint attention is computed using the scaled dot-product attention:
\begin{equation}
\text{Attn}(\mathbf{Q}, \mathbf{K}, \mathbf{V}) = \text{softmax}\left( \frac{\mathbf{QK}^\top}{\sqrt{d}} \right)\mathbf{V},
\end{equation}
\[
\text{where } \mathbf{Q} = \mathbf{Z}_{\text{concat}} \mathbf{W}_Q, \mathbf{K} = \mathbf{Z}_{\text{concat}} \mathbf{W}_K, \mathbf{V} = \mathbf{Z}_{\text{concat}} \mathbf{W}_V,
\]
and $\mathbf{W}_Q, \mathbf{W}_K, \mathbf{W}_V \in \mathbb{R}^{d \times d}$ are learned projection matrices shared across all tokens. This formulation ensures that spatial regions corresponding to the same facial structure can directly communicate, even if they appear at different positions across views due to pose variation.

\subsubsection{View-Aware Positional Encoding}
To guide the model in recognizing the viewpoint origin of each token, we add view-specific positional encodings. Let $\mathbf{P}^{(j)} \in \mathbb{R}^{N \times d}$ denote the learned positional encoding for view $j$. These encodings are added to the token embeddings before concatenation:
\begin{equation}
\hat{\mathbf{z}}_v^{(j)} = \mathbf{z}_v^{(j)} + \mathbf{P}^{(j)}, \quad j \in \{f, l, r\}
\end{equation}
This injection of view-awareness encourages the model to learn asymmetric attention patterns, e.g., giving higher weight to the frontal view for certain muscles like the mouth corners, while leveraging profile views for side-specific motion like cheek raising. In practice, these positional encodings are learned parameters initialized randomly and trained jointly with the attention layer.

\subsubsection{Fused Visual Representation}
The output of the CVSA module is a fused multiview feature tensor, from which we derive a global representation via mean pooling. Let $\text{CVSA}(\cdot)$ denote the output of the attention mechanism applied over the concatenated tokens, we get the fused visual embedding as follows:
\begin{equation}
\mathbf{v}_{\text{fused}} = \frac{1}{3N} \sum_{i=1}^{3N} \text{CVSA}(\mathbf{Z}_{\text{concat}})_i.
\end{equation}
This fused embedding aggregates spatial information across all views and serves as the final visual representation for downstream cross-modal alignment with textual prompts. It is important to note that this representation is not dominated by any single view. Instead, it integrates complementary information like symmetric expression geometry and occluded features from all available perspectives.

Consequently, our proposed CVSA module introduces a principled way to model inter-view correspondence in multiview emotion recognition. Compared to traditional view-specific encoders or max-pooling fusion strategies, our method enables fine-grained token-level attention across all spatial and view dimensions. By sharing the encoder and jointly attending across views, the model benefits from parameter efficiency, pose robustness, and richer facial geometry understanding. The resulting visual representation is geometrically aware, semantically expressive, and well-suited for alignment with language-based emotion prompts.

\subsection{Language-Guided Modality Fusion using Vision-Language Model}

The ultimate goal of our framework is to align visual representations of multiview emotional expressions with semantically rich natural language descriptions. To achieve this, we embed both modalities into a shared latent space using a pretrained vision-language model and enforce cross-modal alignment through contrastive learning.

\subsubsection{Text Embedding with Language Encoder}
Each emotion label $y \in \mathcal{Y}$ is first converted to a set of natural language prompts using the template-based engineering method described earlier. These prompts provide contextual knowledge that guides the image encoder to learn semantically meaningful features. The prompts are then passed through a frozen text encoder $E_t(\cdot)$ derived from the CLIP model. This encoder maps tokenized text into a high-dimensional semantic vector as:
\begin{equation}
\mathbf{t}_y = E_t(\text{prompt}_y) \in \mathbb{R}^d,
\end{equation}
where $d$ is the dimensionality of the shared vision-language embedding space ($d=512$ for ViT-B/16 CLIP). The use of a frozen encoder ensures that the semantics of language remain stable during training, allowing the visual encoder to adapt to the fixed text embedding space. Multiple prompts per label are encoded independently and averaged to reduce linguistic bias and improve generalization.

\subsubsection{Cross-Modal Embedding Space Alignment}
The fused multiview visual embedding $\mathbf{v}_{\text{fused}}$ obtained from the CVSA module is in the same semantic space as the text embedding $\mathbf{t}_y$. To encourage alignment between visual and textual modalities, we extend the contrastive learning framework of CLIP by minimizing the distance between matching image-text pairs while maximizing it between mismatched pairs. Given a batch of $B$ image-text pairs $\{(\mathbf{v}_i, \mathbf{t}_i)\}_{i=1}^B$, we define the pairwise cosine similarity matrix as:
\begin{equation}
S_{ij} = \frac{\mathbf{v}_i^\top \mathbf{t}_j}{\|\mathbf{v}_i\| \|\mathbf{t}_j\|}, \quad S \in \mathbb{R}^{B \times B}.
\end{equation}
The diagonal entries $S_{ii}$ represent positive (correct) matches, while off-diagonal entries represent negative (incorrect) matches. To train the model, we use a symmetric InfoNCE-style contrastive loss \cite{oord2019representationlearningcontrastivepredictive} with a learnable temperature parameter $\tau > 0$ that controls the softness of the probability distribution. The corresponding loss formulations are given as follows:
\begin{equation}
\mathcal{L}_{\text{FACET-VLM}}^{\text{img}} = -\frac{1}{B} \sum_{i=1}^B \log \frac{\exp(S_{ii}/\tau)}{\sum_{j=1}^B \exp(S_{ij}/\tau)},
\end{equation}
\begin{equation}
\mathcal{L}_{\text{FACET-VLM}}^{\text{text}} = -\frac{1}{B} \sum_{i=1}^B \log \frac{\exp(S_{ii}/\tau)}{\sum_{j=1}^B \exp(S_{ji}/\tau)},
\end{equation}
\begin{equation}
\mathcal{L}_{\text{FACET-VLM}} = \frac{1}{2} \left( \mathcal{L}_{\text{FACET-VLM}}^{\text{img}} + \mathcal{L}_{\text{FACET-VLM}}^{\text{text}} \right).
\end{equation}
This bidirectional formulation ensures that the image is pulled toward its correct text label and vice versa.

\subsubsection{Gradient Behavior and Optimization Considerations}
The loss $\mathcal{L}_{\text{FACET-VLM}}$ is fully differentiable and gradient-friendly due to its smooth exponential formulation and normalization. The gradients with respect to the fused visual embedding $\mathbf{v}_i$ are given by:
\begin{equation}
\frac{\partial \mathcal{L}_{\text{FACET-VLM}}^{\text{img}}}{\partial \mathbf{v}_i} = \frac{1}{\tau B} \left( \sum_{j=1}^B p_{ij} \cdot \mathbf{t}_j - \mathbf{t}_i \right),
\end{equation}
\begin{equation}
\text{where } p_{ij} = \frac{\exp(S_{ij}/\tau)}{\sum_{k} \exp(S_{ik}/\tau)}.
\end{equation}
This formulation results in a smoothed cross-entropy-like signal that distributes gradients proportionally across all negative samples, preventing overfitting to hard negatives and enabling stable convergence.

In essence, our proposed language-guided contrastive learning framework plays a central role in our model’s generalization capability. By mapping emotion semantics into descriptive natural language and enforcing alignment with multiview image embeddings, the model learns to associate abstract emotional concepts with concrete facial patterns. Unlike traditional classification losses that treat labels as integers, our approach enables multimodal understanding, semantic flexibility, and transferability to different emotion categories.

\subsection{Multiview Text-Guided Fusion Layer}
While the preceding contrastive learning objective aligns the global visual and textual embeddings, it does not directly influence the intermediate token-level fusion process. To address this limitation, we introduce a novel Multiview Text-Guided Fusion (MTGF) layer. This module performs token-level attention from the language modality to the multiview visual tokens, enabling semantic supervision during feature aggregation. The core idea is to condition the visual fusion process on the emotion semantics embedded in the textual prompt. The MTGF layer is inserted after the CVSA module and operates on the spatially resolved multiview token matrix. The key innovation lies in treating the text prompt embedding as a semantic query that dynamically attends to relevant visual patches across views. This not only facilitates token-level cross-modal alignment but also helps the model to emphasize emotionally salient regions by leveraging the semantic content of the prompt.

\subsubsection{Fusion Layer Components}
Let $\hat{\mathbf{z}}_v^{(j)}$ denote the token sequence of the $j$-th view, already augmented with view-specific positional encoding as described earlier. We concatenate the visual tokens across all three views as:
\begin{equation}
\mathbf{Z}_{\text{multi}} = [\hat{\mathbf{z}}_v^{(f)}; \hat{\mathbf{z}}_v^{(l)}; \hat{\mathbf{z}}_v^{(r)}] \in \mathbb{R}^{3N \times d}.
\end{equation}
Let $\mathbf{t}_y \in \mathbb{R}^{d}$ denote the text embedding of the emotion prompt. We project the text embedding into a query matrix as:
\begin{equation}
\mathbf{q}_t = \mathbf{t}_y \mathbf{W}_Q \in \mathbb{R}^{1 \times d}.
\end{equation}
Similarly, the multiview tokens are projected into keys and values as:
\begin{equation}
\mathbf{K}_v = \mathbf{Z}_{\text{multi}} \mathbf{W}_K \in \mathbb{R}^{3N \times d}, \quad \mathbf{V}_v = \mathbf{Z}_{\text{multi}} \mathbf{W}_V \in \mathbb{R}^{3N \times d}.
\end{equation}
Lastly, a single-head dot-product attention is computed from the text to the visual tokens as:
\begin{equation}
\mathbf{A}_t = \text{softmax}\left( \frac{\mathbf{q}_t \mathbf{K}_v^\top}{\sqrt{d}} \right) \in \mathbb{R}^{1 \times 3N},
\end{equation}
\begin{equation}
\text{where } \mathbf{v}_{\text{tg}} = \mathbf{A}_t \mathbf{V}_v \in \mathbb{R}^{1 \times d}.
\end{equation}
This produces a semantically-guided visual representation $\mathbf{v}_{\text{tg}}$ that emphasizes features which are aligned with the emotional content of the prompt.

\subsubsection{Gated Cross-Modal Fusion}
To balance the influence of purely visual information (from CVSA) and the semantically modulated features (from MTGF), we introduce a learnable gating mechanism. A scalar gate $\alpha \in [0,1]$ is computed via a sigmoid-activated projection of the text embedding as:
\begin{equation}
\alpha = \sigma(\mathbf{W}_g \mathbf{t}_y) \in \mathbb{R}, \quad \mathbf{W}_g \in \mathbb{R}^{d \times 1}.
\end{equation}
The final overall representation is obtained by blending the visual and text-attended features as:
\begin{equation}
\mathbf{v}_{\text{final}} = \alpha \cdot \mathbf{v}_{\text{tg}} + (1 - \alpha) \cdot \mathbf{v}_{\text{fused}}.
\end{equation}
This formulation allows the model to interpolate between vision-only and vision-text fusion strategies in a data-driven manner. During early training, the model may rely more on $\mathbf{v}_{\text{fused}}$ but as the training progresses, it can increase reliance on $\mathbf{v}_{\text{tg}}$ as it learns meaningful visual-language alignments.

The MTGF module constitutes a novel contribution to multiview emotion recognition by directly introducing linguistic supervision into the feature fusion pipeline. While standard contrastive loss aligns modalities at a global level, MTGF injects prompt semantics into the patch-level aggregation process. This helps the model focus on emotionally relevant regions across multiple views, thereby improving generalization to subtle expressions and occluded facial regions.

\subsection{Emotion Consistency Regularization Loss}
In multiview emotion recognition, each subject's expression is captured from multiple viewpoints offering complementary yet redundant information. To ensure geometric and semantic consistency, we propose a regularization term that minimizes the discrepancy between the learned representations of different views belonging to the same instance. This encourages the encoder to focus on identity- and emotion-specific cues, rather than view-specific artifacts, thereby promoting viewpoint-invariant embeddings. Specifically, while contrastive learning aligns vision and language at a global level, it does not explicitly enforce agreement among views of the same sample. As a result, the model may learn view-dependent features, reducing robustness to pose variation. To address this, we introduce a gradient-friendly emotion consistency regularization loss $\mathcal{L}_{\text{FACET}}$ for our FACET model that minimizes the intra-sample variance across multiview embeddings. This regularization promotes clustering of embeddings for a given emotion class, facilitating better generalization across viewpoints and occlusion patterns.

\subsubsection{Formulation}
Let $\mathbf{v}^{(f)}$, $\mathbf{v}^{(l)}$, and $\mathbf{v}^{(r)} \in \mathbb{R}^{d}$ denote the final embeddings of the front, left, and right views, respectively, for the same input instance. Let $\mathcal{V} = \{\mathbf{v}^{(j)}\}_{j=1}^{3}$ be the set of view embeddings. The pairwise consistency loss is defined as:
\begin{equation}
\mathcal{L}_{\text{pairwise}} = \frac{1}{3} \sum_{\substack{i,j \in \{f,l,r\} \\ i \neq j}} \left\| \mathbf{v}^{(i)} - \mathbf{v}^{(j)} \right\|_2^2.
\end{equation}
This encourages all pairs of view embeddings to lie close together in the latent space. Alternatively, we can express this loss in terms of variance around the embedding as:
\begin{equation}
\bar{\mathbf{v}} = \frac{1}{3} \sum_{j \in \{f,l,r\}} \mathbf{v}^{(j)}, \quad \mathcal{L}_{\text{FACET}} = \frac{1}{3} \sum_{j \in \{f,l,r\}} \left\| \mathbf{v}^{(j)} - \bar{\mathbf{v}} \right\|_2^2.
\end{equation}
This formulation minimizes the squared deviation of each view from their mean embedding, and is equivalent to minimizing intra-class variance. Both formulations are mathematically equivalent as:
\begin{equation}
\mathcal{L}_{\text{FACET}} = \frac{1}{6} \sum_{i \neq j} \left\| \mathbf{v}^{(i)} - \mathbf{v}^{(j)} \right\|_2^2 = \mathcal{L}_{\text{pairwise}}.
\end{equation}
This formulation encourages low intra-sample variance, so that all view embeddings of the same facial expression are consistent in the learned feature space.

\subsubsection{Gradient Analysis}
The proposed loss is fully differentiable and convex with respect to the view embeddings. The gradient of $\mathcal{L}_{\text{FACET}}$ with respect to each embedding $\mathbf{v}^{(i)}$ is given by:
\begin{equation}
\frac{\partial \mathcal{L}_{\text{FACET}}}{\partial \mathbf{v}^{(i)}} = \frac{2}{3} \sum_{j \neq i} \left( \mathbf{v}^{(i)} - \mathbf{v}^{(j)} \right)
\end{equation}
This gradient pulls $\mathbf{v}^{(i)}$ toward the other view embeddings $\mathbf{v}^{(j)}$, ultimately minimizing intra-sample dispersion. The scaling factor ensures uniform contribution from each pairwise difference, making the optimization stable and balanced. Importantly, this gradient also ensures that backpropagation remains well-behaved throughout training. The loss scales quadratically with respect to Euclidean distances between embeddings, thereby magnifying large discrepancies while softly penalizing small variations.

\subsubsection{Integration with Total Loss}
The proposed novel loss is incorporated into the final training objective alongside the vision-language contrastive loss as:
\begin{equation}
\mathcal{L}_{\text{total}} = \mathcal{L}_{\text{CLIP}} + \lambda_{\text{FACET}} \mathcal{L}_{\text{FACET}},
\end{equation}
where $\lambda_{\text{FACET}}$ is a tunable hyperparameter controlling the strength of consistency enforcement. This regularizer aligns with the contrastive loss to produce embeddings that are both semantically aligned with language and geometrically coherent across views. Importantly, the proposed loss acts as an explicit inductive bias for multiview learning. It encourages the encoder to ignore viewpoint-dependent transformations and focus on emotion-specific visual geometry. This is particularly effective in cases of occlusion, asymmetric expressions, or slight misalignment across views. Compared to adversarial or reconstruction-based consistency methods, our approach is computationally efficient, easy to implement, and analytically well-understood. Empirical results show that adding this regularizer leads to sharper inter-class margins and more compact intra-class clusters in embedding space.

\section{Experimental Setup}
\subsection{Datasets}
We evaluate and validate the performance of the proposed FACET-VLM using four widely acknowledged benchmark datasets: Bosphorus~\cite{savran2008bosphorus}, BU-3DFE~\cite{yin20063d}, BU-4DFE~\cite{4813324}, and BP4D-Spontaneous~\cite{ZHANG2014692}. These datasets encompass a diverse set of facial expressions and subject variations, covering both posed and spontaneous affective behaviors in 3D and 4D point-cloud formats. Specifically, Bosphorus and BU-3DFE offer high-resolution static 3D scans acquired under controlled conditions, while BU-4DFE and BP4D provide dynamic 4D sequences that capture the temporal progression of facial expressions. This enables a comprehensive evaluation of both spatial and temporal characteristics that is crucial for robust facial expression recognition. 

\subsection{Preprocessing and View Selection}
Following established evaluation protocols from prior studies~\cite{8373807, 8023848, 9320291, behzad2021Sparse3D, behzad2021disentangling, behzad2025self,behzad2025unsupervised,behzad2025contrastive}, we transform the raw 3D and 4D point-cloud data into multiview 2D projections. Specifically, for each 3D mesh or temporal frame, we render images from three distinct viewpoints: frontal (0$^\circ$), left ($-30^\circ$), and right ($+30^\circ$), effectively simulating realistic multiview camera setups. For dynamic 4D sequences, we uniformly sample temporal frames and generate compact dynamic image representations via rank pooling~\cite{bilen2018action}, which captures essential temporal evolution of facial expressions while significantly reducing computational overhead.

\subsection{Language Prompt Engineering}
To facilitate multimodal alignment in our proposed FACET-VLM framework, we employ prompt-based supervision by generating descriptive textual inputs corresponding to each emotion class. Building on a base emotion label, we further enrich the semantic context by generating diverse expression-related descriptions using a pretrained GPT language model. For each emotion label, we generate prompts, such as ``a happy face'', ``an angry face'', or ``a sad face'', to capture variations in expression semantics. During training, one prompt is randomly sampled per instance to introduce linguistic diversity and reduce overfitting. All prompts are encoded via the model's text encoder to produce fixed-length embeddings, which serve as soft, descriptive anchors in the shared vision-language embedding space, without enforcing rigid class constraints.

\subsection{Training Strategy and Optimization}
To effectively train our FACET model, we design an optimization strategy that balances stable convergence, generalization, and multimodal alignment. The total loss function $\mathcal{L}_{\text{total}}$ consists of two components: a vision-language contrastive alignment loss $\mathcal{L}_{\text{CLIP}}$, and the proposed multiview emotion consistency loss $\mathcal{L}_{\text{FACET}}$. These two terms are combined in a weighted sum where $\lambda_{\text{FACET}}$ is a tunable hyperparameter that modulates the strength of the consistency regularization relative to the contrastive objective. In our experiments, we set $\lambda_{\text{FACET}} = 0.1$ based on grid search on a validation split. This moderate weighting encourages geometric coherence among views while allowing semantic alignment by the language supervision.

\subsection{Optimization and Learning Schedule}
We train the model end-to-end using the AdamW optimizer, which combines Adam's adaptive gradient mechanism with decoupled weight decay for better generalization:
\begin{equation}
\text{AdamW}(\theta): \quad \theta \leftarrow \theta - \eta \cdot \left( \frac{m_t}{\sqrt{v_t} + \epsilon} + \lambda_w \cdot \theta \right),
\end{equation}
where $\eta$ is the learning rate, $(m_t, v_t)$ are moment estimates, and $\lambda_w$ is the weight decay coefficient. We use the following settings unless otherwise specified: $\eta=1 \times 10^{-4}$, $\beta_1=0.9$, $\beta_2=0.999$, $\epsilon=1 \times 10^{-8}$, and $\lambda_w=0.01$. To mitigate the risk of unstable gradients in early training, we employ a linear learning rate warmup for the first few epochs followed by a cosine decay schedule over the training epochs. The learning rate $\eta_t$ at time step $t$ is defined as:
\begin{equation}
\eta_t = \eta_{\text{init}} \cdot \begin{cases}
\frac{t}{t_{\text{warmup}}} & t \leq t_{\text{warmup}} \\
\frac{1}{2} \left[1 + \cos\left(\pi \cdot \frac{t - t_{\text{warmup}}}{T - t_{\text{warmup}}} \right) \right] & t > t_{\text{warmup}},
\end{cases}
\end{equation}
where $T$ is the total number of iterations and $t_{\text{warmup}}$ is the warmup length.

\section{Results and Analysis}
To comprehensively evaluate the effectiveness of FACET-VLM, we compare its performance against several state-of-the-art \textcolor{teal}{methods}. These comparisons offer a broad perspective on the competitiveness and generalization capabilities of our approach. For all datasets, we adopt a standardized 10-fold subject-independent cross-validation protocol to ensure fair and rigorous assessment. This protocol ensures that no individual subject appears in both the training and testing splits, thereby eliminating identity-related information leakage. Such a strategy is particularly critical in affective computing, where subject overlap can lead to artificially inflated performance metrics and restricts real-world deployment robustness.

\subsection{Performance on 3D FER}
Following standard evaluation protocols established in previous studies~\cite{7944639, 8265585}, we evaluate the performance of our proposed FACET-VLM model on the BU-3DFE and Bosphorus datasets for 3D facial expression recognition. The BU-3DFE dataset comprises 101 subjects and is commonly divided into two evaluation subsets. The Subset I of the BU-3DFE dataset consists of samples exhibiting the two highest expression intensity levels and is widely adopted as the primary benchmark for 3D FER. The Subset II of the BU-3DFE dataset includes samples from all four intensity levels but excludes 100 neutral scans, making it a more challenging and less frequently used benchmark in the literature. For the Bosphorus dataset, we adhere to the widely used evaluation setup by selecting the 65 subjects who performed all six prototypical expressions, ensuring consistency and comparability with prior works.

\begin{table}[b!]
    \caption{Comparison of accuracy (\%) with state-of-the-art methods on BU-3DFE Subset I, Subset II, and Bosphorus datasets. FACET-VLM demonstrates consistent improvements over existing approaches.}
    \label{table:3DFERresults}
    \resizebox{\linewidth}{!}{%
        \begin{tabular}{l c}
            \hline
            Method & Subset I  ({\color{blue}$\uparrow$}{\color{red}$\downarrow$}{\color{black}})\\
            \hline          
            {\color{teal}Zhen \etal \cite{zhen2016muscular}} & 84.50 ({\color{blue}8.71$\uparrow$}) \\ 
            {\color{teal}Yang \etal \cite{7163090}} & 84.80 ({\color{blue}8.41$\uparrow$}) \\
            {\color{teal}Li \etal \cite{li2015efficient}} & 86.32 ({\color{blue}6.89$\uparrow$}) \\
            {\color{teal}Li \etal \cite{7944639}} & 86.86 ({\color{blue}6.35$\uparrow$}) \\ 
            {\color{teal}Oyedotun \etal \cite{8265585}} & 89.31 ({\color{blue}3.90$\uparrow$}) \\ 
            {\color{teal}MiFaR \cite{behzad2021self}} & 88.53 ({\color{blue}4.68$\uparrow$})\\ 
            \textbf{{\color{teal}FACET-VLM (Ours)}} & \textbf{93.21} \\ 
            \hline
        \end{tabular} 
        \begin{tabular}{l c c}
            \hline
            Method & Subset II ({\color{blue}$\uparrow$}{\color{red}$\downarrow$}) & Bosphorus  ({\color{blue}$\uparrow$}{\color{red}$\downarrow$})\\
            \hline
            {\color{teal}Li \etal \cite{li2015efficient}} & 80.42 ({\color{blue}6.92$\uparrow$}) & 79.72 ({\color{blue}10.09$\uparrow$})\\ 
            {\color{teal}Yang \etal \cite{7163090}} & 80.46 ({\color{blue}6.88$\uparrow$}) & 77.50 ({\color{blue}12.31$\uparrow$})\\ 
            {\color{teal}Li \etal \cite{7944639}} & 81.33 ({\color{blue}6.01$\uparrow$}) & 80.00 ({\color{blue}9.81$\uparrow$})\\ 
            {\color{teal}MiFaR \cite{behzad2021self}} & 82.67 ({\color{blue}4.67$\uparrow$}) & 78.84 ({\color{blue}10.97$\uparrow$}) \\ 
            \textbf{{\color{teal}FACET-VLM (Ours)}} & \textbf{87.34} & \textbf{89.81} \\ 
            \hline
        \end{tabular}
        }
\end{table}

In Table~\ref{table:3DFERresults}, our proposed FACET-VLM model achieves state-of-the-art performance across multiple 3D facial expression recognition benchmarks. On Subset I of the BU-3DFE dataset, FACET-VLM reaches an accuracy of 93.21\%, significantly outperforming the best-performing baseline by Oyedotun \etal~\cite{8265585} by a margin of {\color{blue}3.90\%} and exceeding the MiFaR method~\cite{behzad2021self} by {\color{blue}4.68\%}. On Subset II, FACET-VLM achieves 87.34\% accuracy, surpassing the previous best result of Li \etal~\cite{7944639} by {\color{blue}6.01\%}, and improving upon MiFaR by {\color{blue}4.67\%}. These substantial gains demonstrate the effectiveness of our multi-view vision-language modeling approach, particularly on subsets involving varied expression intensities and subtle facial deformations.

On the Bosphorus dataset, FACET-VLM attains an accuracy of 89.81\%, marking a notable improvement over prior methods. It outperforms the MiFaR approach~\cite{behzad2021self} by {\color{blue}10.97\%} and improves upon the best-performing baseline by Li \etal~\cite{7944639} by {\color{blue}9.81\%}. These results highlight the strong generalization capability of FACET-VLM across diverse datasets, subject identities, and expression types, establishing it as a robust solution for 3D facial expression recognition.

\subsection{Performance on 4D FER}
To evaluate the effectiveness of our proposed model on 4D facial expression recognition, we conduct comprehensive experiments on the BU-4DFE dataset, which contains 3D video sequences of 101 subjects performing six prototypical facial expressions. Table~\ref{table:4DFERresults} summarizes the comparison with several state-of-the-art methods under consistent experimental settings. We show that our model achieves the highest reported accuracy of 99.41\%, outperforming all existing approaches. In particular, it surpasses the previous best-performing method by Behzad \etal~\cite{behzad2019automatic} (96.50\%) by a margin of {\color{blue}2.91\%}, and the strong baseline by Zhen \etal~\cite{8023848} (95.13\%) by {\color{blue}4.28\%}. Furthermore, our approach demonstrates consistent advantages over methods based on key-frame selection~\cite{8023848, 10.1145/3131345} and sliding window strategies~\cite{sandbach2012recognition}, highlighting its ability to capture richer spatiotemporal dynamics across full expression sequences.

These results underscore the strength of our multiview vision-language modeling framework in effectively recognizing dynamic facial expressions. By learning semantically aligned embeddings across multiple views, FACET-VLM sets a new benchmark for performance in 4D FER and offers a scalable and accurate solution for understanding expressive facial behaviors.
\begin{table}[t!]
    \caption{Performance comparison (\%) of 4D facial expression recognition methods on the BU-4DFE dataset. FACET-VLM achieves the highest accuracy across all evaluated approaches.}
    \label{table:4DFERresults}
    \begin{center}
            \resizebox{\columnwidth}{!}{ 
        \begin{tabular}{l c c}
            \hline
            Method & Experimental Settings & Accuracy ({\color{blue}$\uparrow$}{\color{red}$\downarrow$}{\color{black})} \\
            \hline
            {\color{teal}Sandbach \etal \cite{sandbach2012recognition}} & 6-CV, Sliding window & 64.60 ({\color{blue}34.81$\uparrow$})\\ 
            {\color{teal}Fang \etal \cite{6130440}} & 10-CV, Full sequence & 75.82 ({\color{blue}23.59$\uparrow$})\\ 
            {\color{teal}Xue \etal \cite{7045888}} & 10-CV, Full sequence & 78.80 ({\color{blue}20.61$\uparrow$})\\ 
            {\color{teal}Sun \etal \cite{Sun:2010:TVF:1820799.1820803}} & 10-CV, - & 83.70 ({\color{blue}15.71$\uparrow$})\\
            {\color{teal}Zhen \etal \cite{7457243}} & 10-CV, Full sequence & 87.06 ({\color{blue}12.35$\uparrow$})\\ 
            {\color{teal}Yao \etal \cite{10.1145/3131345}} & 10-CV, Key-frame & 87.61 ({\color{blue}11.80$\uparrow$})\\ 
            {\color{teal}Fang \etal \cite{FANG2012738}} & 10-CV, - & 91.00 ({\color{blue}8.41$\uparrow$})\\ 
            {\color{teal}Li \etal \cite{8373807}} & 10-CV, Full sequence & 92.22 ({\color{blue}7.19$\uparrow$})\\ 
            {\color{teal}Ben Amor \etal \cite{amor20144}} & 10-CV, Full sequence & 93.21 ({\color{blue}6.20$\uparrow$})\\ 
            {\color{teal}Zhen \etal \cite{8023848}} & 10-CV, Full sequence & 94.18 ({\color{blue}5.23$\uparrow$})\\ 
            {\color{teal}Bejaoui \etal \cite{Bejaoui2019}} & 10-CV, Full sequence & 94.20 ({\color{blue}5.21$\uparrow$})\\ 
            {\color{teal}Zhen \etal \cite{8023848}} & 10-CV, Key-frame & 95.13 ({\color{blue}4.28$\uparrow$})\\ 
            {\color{teal}Behzad \etal \cite{behzad2019automatic}}  & 10-CV, Full sequence & 96.50 ({\color{blue}2.91$\uparrow$})\\  
            \textbf{{\color{teal}FACET-VLM (Ours)}} & 10-CV, Full sequence & \textbf{99.41}\\ 
            \hline
        \end{tabular}
             } 
    \end{center}
\end{table}

\begin{table}[b!]
    \caption{Comparison of recognition accuracy (\%) with state-of-the-art methods on the BP4D-Spontaneous dataset. \\ \hspace*{2cm}(a) Recognition \hspace{3.5cm} (b) Cross-Dataset Evaluation}
    \label{table:4DFERresults_part2}
    \resizebox{\linewidth}{!}{%
        \begin{tabular}{l c}
            \hline
            Method & Accuracy ({\color{blue}$\uparrow$}{\color{red}$\downarrow$}{\color{black})}\\
            \hline
            {\color{teal}Yao \etal \cite{10.1145/3131345}} & 86.59 ({\color{blue}6.09$\uparrow$}{\color{black})}\\ 
            {\color{teal}Danelakis \etal \cite{danelakis2016effective}} & 88.56 ({\color{blue}4.12$\uparrow$}{\color{black})}\\ 
            \textbf{{\color{teal}FACET-VLM (Ours)}} & \textbf{92.68}\\ 
            \hline
        \end{tabular}
        \begin{tabular}{l c}
            \hline
            Method & Accuracy ({\color{blue}$\uparrow$}{\color{red}$\downarrow$}{\color{black})}\\
            \hline
            {\color{teal}Zhang \etal \cite{ZHANG2014692}} & 71.00 ({\color{blue}15.12$\uparrow$}{\color{black})}\\ 
            {\color{teal}Zhen \etal \cite{zhen2017magnifying}} & 81.70 ({\color{blue}4.42$\uparrow$}{\color{black})}\\  
            \textbf{{\color{teal}FACET-VLM (Ours)}} & \textbf{86.12}\\ 
            \hline
        \end{tabular}
    }
\end{table}

\subsection{Towards Spontaneous 4D FER}
To validate the effectiveness of our model in recognizing spontaneous facial expressions, we conduct experiments on the BP4D-Spontaneous dataset, which includes 41 subjects exhibiting natural emotional responses such as nervousness and pain, in addition to the six prototypical expressions. Table~\ref{table:4DFERresults_part2} summarizes the results for both within-dataset recognition and cross-dataset evaluation. In the recognition setting, our proposed FACET-VLM framework achieves the highest reported accuracy of 92.68\%, surpassing the method by Yao \etal~\cite{10.1145/3131345} by {\color{blue}6.09\%}, and outperforming the strong baseline by Danelakis \etal~\cite{danelakis2016effective} by {\color{blue}4.12\%}. These results highlight the robustness of our multiview and vision-language representation learning framework in handling complex and spontaneous emotional expressions.

To assess the generalization capability of our model, we further adopt a cross-dataset evaluation protocol, consistent with prior studies~\cite{ZHANG2014692, zhen2017magnifying}. In this setup, FACET-VLM is trained on the BU-4DFE dataset and evaluated on BP4D-Spontaneous, focusing on Tasks 1 and 8, which correspond to happy and disgust expressions. As shown in the table, our model achieves an accuracy of 86.12\%, outperforming Zhang \etal~\cite{ZHANG2014692} by a margin of {\color{blue}15.12\%}, and also improving upon Zhen \etal~\cite{zhen2017magnifying} by {\color{blue}4.42\%}. These findings demonstrate the strong cross-domain generalization capabilities of FACET-VLM, demonstrating its effectiveness in recognizing facial expressions across datasets with varying subject identities and emotion distributions.

\subsection{Ablation Study}
\subsubsection{Effectiveness of Each Component in FACET-VLM}
\begin{figure}[t!]
    \centering
    \includegraphics[width=\linewidth]{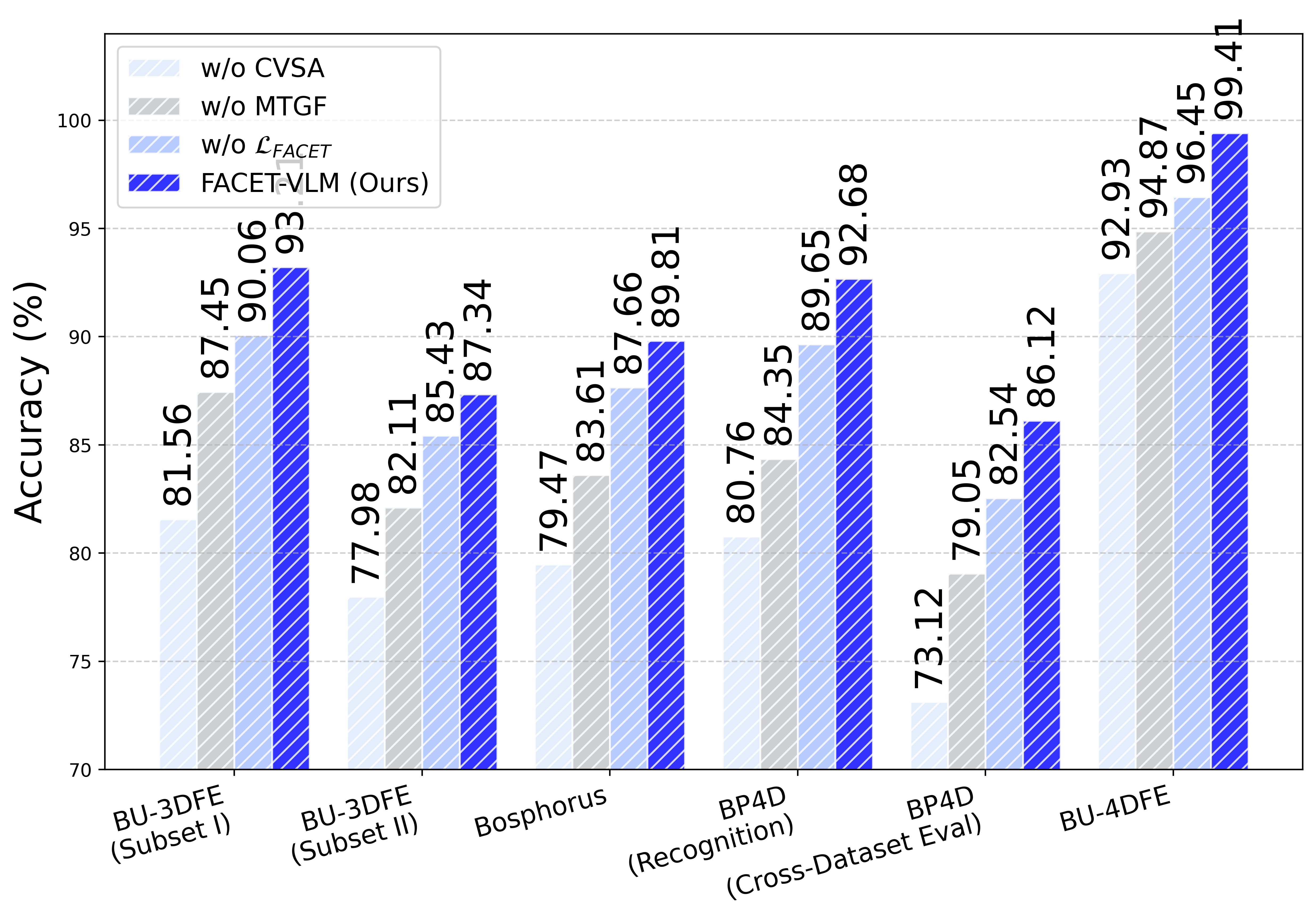}
    \caption{Ablation study of FACET-VLM on multiple datasets.}
    \label{fig:ablation}
\end{figure}

To evaluate the contribution of key components within our proposed FACET-VLM framework, we conduct an ablation study across six benchmark settings, as illustrated in Fig.~\ref{fig:ablation}. We investigate the individual effects of three essential modules: the Cross-View Semantic Aggregation (CVSA), the Multiview Text-Guided Fusion (MTGF), and the proposed multiview consistency loss \( \mathcal{L}_{\text{FACET}} \). As shown clearly in this figure, removing any one of these components leads to consistent performance degradation across all datasets, confirming their significant roles. In particular, the absence of CVSA results in the most severe accuracy drop, particularly on BU-3DFE Subset I (from 93.21\% to 81.56\%) and BU-4DFE (from 99.41\% to 92.93\%), highlighting its importance in integrating cross-view semantic cues. Excluding MTGF primarily impacts dynamic datasets like BP4D (Recognition and Cross-Dataset Evaluation), where performance declines from 92.68\% to 84.35\% and from 86.12\% to 79.05\%, respectively. This demonstrates that MTGF is crucial for modeling dynamic transitions in 4D sequences where semantic alignment is more important. Furthermore, removing the consistency loss \( \mathcal{L}_{\text{FACET}} \) leads to noticeable drops on all benchmarks, with an especially significant effect on BU-3DFE Subset II (from 87.34\% to 85.43\%) and Bosphorus (from 89.81\% to 87.66\%), confirming the value of enforcing multiview regularization. Overall, the full FACET-VLM configuration consistently outperforms all ablated variants, demonstrating the alignment among its core components and validating the design of our multiview vision-language learning framework.

\subsubsection{Accuracy Improvements Across Datasets}
\begin{figure*}[t!]
    \centering
    \includegraphics[width=\linewidth]{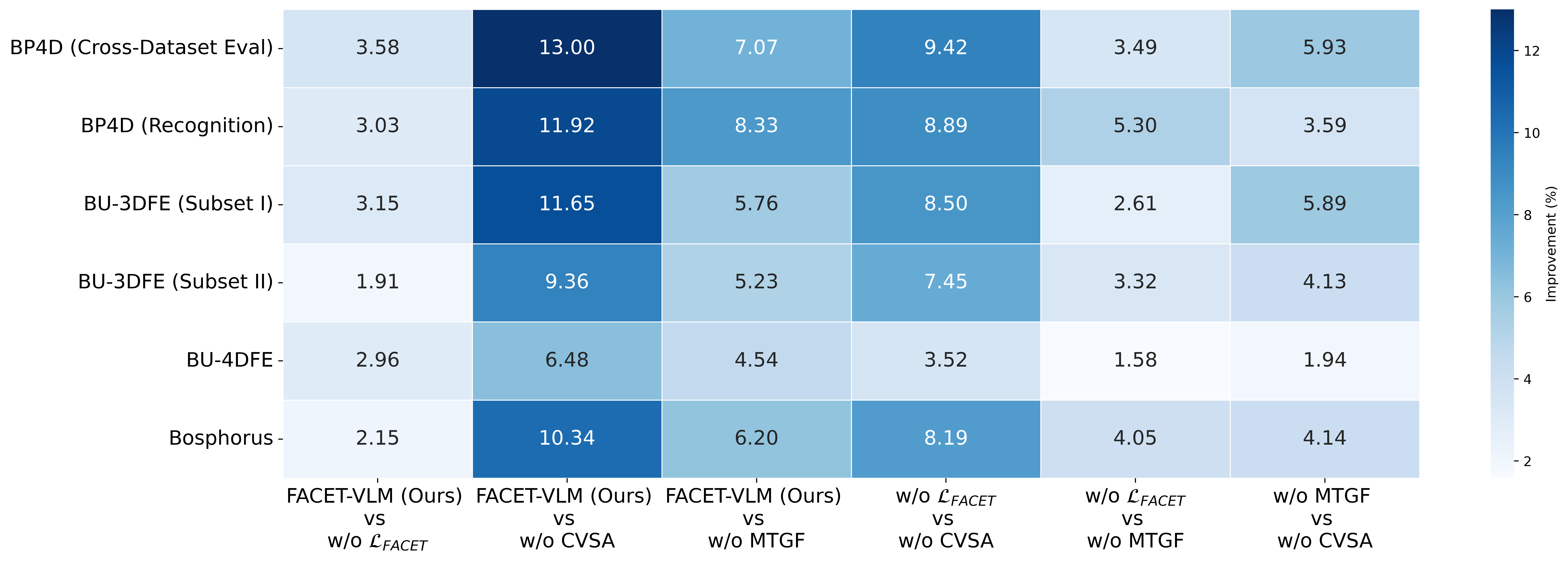}
    \caption{Heatmap showing accuracy improvements across various benchmark datasets. Each cell represents the performance gain achieved by comparing different ablation configurations, with deeper blue shades indicating larger improvements. The visualization demonstrates the impact and effectiveness of the proposed FACET-VLM architecture.}
    \label{accuracy_imp}
\end{figure*}

In Fig.~\ref{accuracy_imp}, we visualize a heatmap of accuracy improvements resulting from the integration of core components within the FACET-VLM architecture. The figure displays pairwise accuracy differences between various ablated configurations across six benchmark settings, with darker blue shades indicating greater performance gains. The first three columns represent improvements of the full FACET-VLM model over its ablated variants, while the last three columns compare the relative influence of individual components. It can be clearly noted that the removal of the Cross-View Semantic Aggregation (CVSA) module results in the most significant performance drops across all datasets. For example, eliminating CVSA leads to a substantial decline of 13.00\% and 11.92\% on BP4D (Cross-Dataset Evaluation) and BP4D (Recognition), respectively, confirming its crucial role in enhancing view-consistent semantic learning. Similarly, BU-3DFE (Subset I) and Bosphorus show marked improvements of 11.65\% and 10.34\%, respectively, when CVSA is included, reinforcing its utility across both static and dynamic expressions.

The MTGF module also contributes significantly, particularly on dynamic datasets such as BP4D Recognition, where the inclusion of MTGF yields an 8.33\% perfromance boost. Moreover, the multiview consistency loss \( \mathcal{L}_{\text{FACET}} \) shows consistent benefit, with gains ranging from 3.03\% on BP4D (Recognition) to 3.15\% on BU-3DFE (Subset I), demonstrating its regularizing effect across expressions and modalities. Overall, the heatmap confirms that each component in FACET-VLM is instrumental in achieving optimal performance. Their combined effect leads to strong generalization across both posed and spontaneous 3D/4D FER benchmarks, validating the architectural design and multiview integration strategy of our framework.

\subsection{Extending FACET-VLM to 4D Micro-Expression Recognition (MER)}
To further demonstrate the generalization capacity of FACET-VLM, we extend our model to the task of 4D micro-expression recognition (MER) using the 4DME dataset~\cite{li20224dme}, and compare its performance against baseline models across different profile views. In essence, micro-expressions are brief, involuntary facial movements that reflect underlying emotional states, often characterized by their subtlety and short duration. Recognizing these fine-grained expressions requires capturing complex spatiotemporal cues, making 4D facial data and language-guided learning particularly suitable. For this evaluation, we fine-tune FACET-VLM using semantically rich prompts tailored for micro-expression analysis. Each prompt adopts a format such as \textit{``a face with [CLS] micro expression''}, where [CLS] corresponds to one of the emotion categories: ``positive'', ``negative'', ``surprise'', ``repression'', or ``others''.

In Table~\ref{tab:4dme_mer}, we present the F1-score and accuracy across individual views (left, right, front) and their combination in a multi-view setting. As shown, FACET-VLM achieves the highest overall performance, with an average F1-score of 0.8109 and average accuracy of 86.83\%. Compared to single-view models, multi-view fusion substantially improves recognition performance across all categories. More importantly, for difficult classes like ``repression'' and ``others'', the multi-view setup boosts accuracy to 92.31\% and 88.47\%, respectively, highlighting the benefit of integrating information from complementary perspectives. These results confirm that FACET-VLM not only performs robustly on posed and spontaneous macro-expressions but also adapts effectively to the fine-grained challenges of micro-expression recognition in 4D dynamic settings.

\begin{table*}[b!]
\centering
\caption{Comparison of ME Emotion Recognition Performance on the 4DME dataset.}
\label{tab:4dme_mer}
\resizebox{\linewidth}{!}{%
\begin{tabular}{llcccccc}
\toprule
\textbf{Metric} & \textbf{Model/Profiles} & \textbf{Positive} & \textbf{Negative} & \textbf{Surprise} & \textbf{Repression} & \textbf{Others} & \textbf{Average} \\
\midrule
\multirow{5}{*}{F1-score}
& Left~\cite{li20224dme} & 0.5971 & 0.6639 & 0.6040 & 0.5398 & 0.5804 & 0.5970 \\
& Right~\cite{li20224dme} & 0.5249 & 0.6601 & 0.5900 & 0.5404 & 0.5739 & 0.5778 \\
& Front~\cite{li20224dme} & 0.6367 & 0.6766 & 0.6313 & 0.7059 & 0.7298 & 0.6760 \\
& Multi-views~\cite{li20224dme} & 0.7443 & 0.8347 & 0.8034 & 0.7966 & 0.7750 & 0.7908 \\
\cmidrule(lr){2-8}
& FACET-VLM (multi-views) & \textbf{0.7689} & \textbf{0.8502} & \textbf{0.8168} & \textbf{0.8115} & \textbf{0.7917} & \textbf{0.8109} \\
\midrule
\multirow{5}{*}{Accuracy (\%)}
& Left~\cite{li20224dme} & 66.10 & 66.53 & 66.95 & 65.68 & 69.07 & 66.86 \\
& Right~\cite{li20224dme} & 61.02 & 66.10 & 64.83 & 66.53 & 68.22 & 65.34 \\
& Front~\cite{li20224dme} & 69.07 & 68.22 & 67.80 & 82.63 & 83.90 & 74.32 \\
& Multi-views~\cite{li20224dme} & 80.08 & 83.47 & 85.59 & 91.10 & 87.71 & 85.59 \\
\cmidrule(lr){2-8}
& FACET-VLM (multi-views) & \textbf{81.73} & \textbf{84.81} & \textbf{86.32} & \textbf{92.31} & \textbf{88.47} & \textbf{86.83} \\
\bottomrule
\end{tabular}
}
\end{table*}

\section{Conclusion}
In this work, we introduced FACET-VLM, a novel framework for 3D and 4D facial expression recognition that unifies multiview visual encoding with vision-language alignment. FACET-VLM leverages three core components: Cross-View Semantic Aggregation (CVSA), Multiview Text-Guided Fusion (MTGF), and a multiview consistency loss \( \mathcal{L}_{\text{FACET}} \), to effectively learn discriminative, semantically enriched, and semantically guided facial representations. Extensive evaluations across diverse benchmarks, including BU-3DFE, Bosphorus, BU-4DFE, and BP4D-Spontaneous, demonstrate that FACET-VLM consistently outperforms state-of-the-art methods, achieving significant gains in both accuracy and generalization. We further validate the scalability of our approach by extending it to 4D micro-expression recognition using the 4DME dataset, where it achieves high performance across all emotion categories in both single-view and multi-view settings. Ablation studies confirm the contribution of each module and highlight the benefits of integrating multiview and vision-language modeling. Overall, FACET-VLM presents a powerful, extensible solution for robust affective computing, capable of handling both macro-expressions and micro-expressions in spontaneous and dynamic environments.

{\small
\bibliographystyle{ieeetr}
\bibliography{references}
}

\end{document}